\documentclass[sigconf]{acmart}

\copyrightyear{2025}
\acmYear{2025}
\setcopyright{acmlicensed}
\acmConference[MM '25] {Proceedings of the 33rd ACM International Conference on Multimedia}{October 27--31, 2025}{Dublin, Ireland.}
\acmBooktitle{Proceedings of the 33rd ACM International Conference on Multimedia (MM '25), October 27--31, 2025, Dublin, Ireland}
\acmISBN{979-8-4007-2035-2/2025/10}
\acmDOI{10.1145/3746027.3758213}

\usepackage[normalem]{ulem}
\usepackage{enumitem}
\usepackage{makecell}
\usepackage{multirow} 

\begin{document}

\title{AU-IQA: A Benchmark Dataset for Perceptual Quality Assessment of AI-Enhanced User-Generated Content}

\author{Shushi Wang}
\affiliation{%
  \institution{Shanghai Jiao Tong University}
  \city{Shanghai}
  \country{China}
}
\email{wss2002@sjtu.edu.cn}

\author{Chunyi Li}
\affiliation{%
 \institution{Shanghai Jiao Tong University}
  \city{Shanghai}
  \country{China}
}
\email{lcysyzxdxc@sjtu.edu.cn}

\author{Zicheng Zhang}
\affiliation{%
  \institution{Shanghai Jiao Tong University}
  \city{Shanghai}
  \country{China}
  }
\email{zzc1998@sjtu.edu.cn}

\author{Han Zhou}
\affiliation{%
  \institution{McMaster University}
  \city{Hamilton}
  \country{Canada}
  }
\email{zhouh115@mcmaster.ca}

\author{Wei Dong}
\affiliation{%
  \institution{McMaster University}
  \city{Hamilton}
  \country{Canada}
  }
\email{dongw22@mcmaster.ca}

\author{Jun Chen}
\affiliation{%
  \institution{McMaster University}
  \city{Hamilton}
  \country{Canada}
  }
\email{chenjun@mcmaster.ca}

\author{Guangtao Zhai}
\affiliation{%
  \institution{Shanghai Jiao Tong University}
  \city{Shanghai}
  \country{China}}
\email{zhaiguangtao@sjtu.edu.cn}

\author{Xiaohong Liu}
\authornote{Corresponding author.}
\affiliation{
  \institution{Shanghai Jiao Tong University}
  \city{Shanghai}
  \country{China}
}
\affiliation{
  \institution{\mbox{\hspace*{-1.5em}Suzhou Key Laboratory of Artificial Intelligence}}
  \city{Suzhou}
  \country{China}
}

\email{xiaohongliu@sjtu.edu.cn}

\renewcommand{\shortauthors}{Shushi Wang et al.}

\begin{abstract}
AI-based image enhancement techniques have been widely adopted in various visual applications, significantly improving the perceptual quality of user-generated content (UGC). However, the lack of specialized quality assessment models has become a significant limiting factor in this field, limiting user experience and hindering the advancement of enhancement methods. While perceptual quality assessment methods have shown strong performance on UGC and AIGC individually, their effectiveness on AI-enhanced UGC (AI-UGC) which blends features from both—remains largely unexplored. To address this gap, we construct AU-IQA, a benchmark dataset comprising 4,800 AI-UGC images produced by three representative enhancement types which include super-resolution, low-light enhancement, and denoising. On this dataset, we further evaluate a range of existing quality assessment models, including traditional IQA methods and large multimodal models. Finally, we provide a comprehensive analysis of how well current approaches perform in assessing the perceptual quality of AI-UGC. The access link to the AU-IQA is \url{https://github.com/WNNGGU/AU-IQA-Dataset}. 
\end{abstract}


\begin{teaserfigure}
  \includegraphics[width=\textwidth]{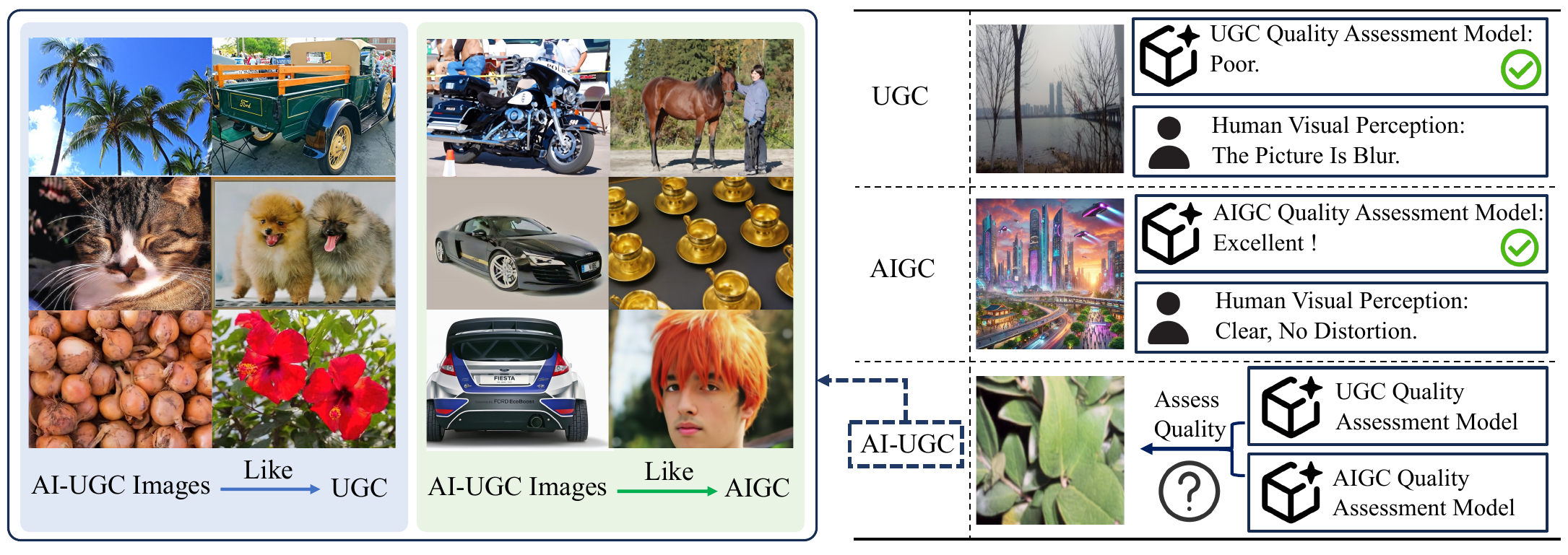}
  \caption{Research significance of AI-UGC quality assessment in practical applications.}
  \label{fig:First_Image}
\end{teaserfigure}
\maketitle




\begin{figure*}[htbp]
\captionsetup{aboveskip=5pt, belowskip=0pt} 
\hspace{2mm}
\centerline{\hspace*{0.15cm}\includegraphics[scale=0.62]{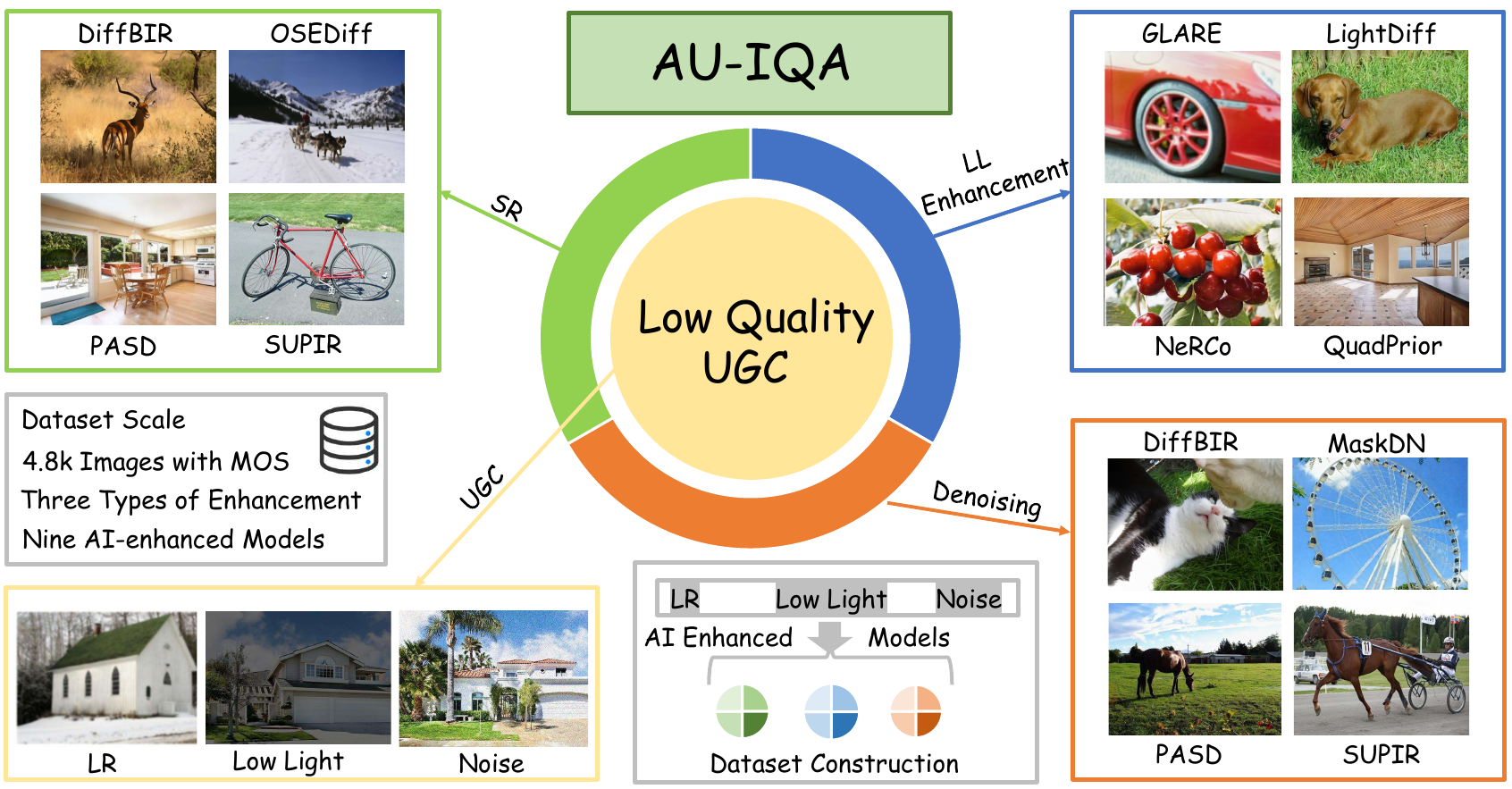}} 
\caption{Our proposed AI-UGC quality assessment dataset, AU-IQA. The dataset consists of three types of AI-UGC generated by nine different AI-enhanced models, totaling 4.8k images.}
\vspace{-0.cm} %
\label{fig_overall}
\end{figure*}
\section{Introduction}

The rapid advancement of generative models ~\cite{goodfellow2014generative, kingma2013auto, ramesh2022hierarchical, rombach2022high} and AI-based enhancement techniques ~\cite{ECMamba, Han_ECCV24_GLARE, Jiang_2024_ECCV, dong2024lightvqa, shi2024video, Attentionlut, hu2025VARFVV} has led to the widespread emergence of AI-enhanced user-generated content (AI-UGC), such as images that have been super-resolved, denoised, deblurred, or otherwise altered using machine learning algorithms. On multimedia platforms such as Instagram, TikTok, and YouTube, AI-UGC is increasingly common. However, despite their growing popularity, AI-UGC can often suffer from artifacts, over-smoothing, unnatural textures, or hallucinated details—issues that may degrade user experience rather than improve it. As a result, evaluating the perceptual quality of such AI-enhanced content has become an urgent and under-explored problem in both academia and industry.

While AI-enhanced user-generated content (AI-UGC) is becoming increasingly prevalent across multimedia platforms, the methods commonly used to assess its quality still heavily rely on pixel-level metrics such as Peak Signal-to-Noise Ratio (PSNR) and Structural Similarity Index Measure (SSIM). These metrics, while useful for detecting low-level distortions, are insufficient for evaluating the perceptual quality of images that have undergone complex enhancement processes, such as super-resolution, denoising, or generative retouching. In contrast, the field of perceptual quality assessment, designed to align with the human visual perceptual system, has made substantial progress in both traditional user-generated content (UGC) and AI-generated content (AIGC). For these domains, a variety of benchmark datasets ~\cite{hosu2020koniq, li2023agiqa, li2024aigiqa, kou2024Subjective-Aligned, zhou2024thqa, zhang2024metric, TowardsOpen-ended} and learning-based IQA models ~\cite{zhang2023blind, wang2024large} have been developed, focusing on capturing perceptual quality rather than relying solely on pixel-wise accuracy.

Given that AI-UGC simultaneously inherits characteristics from both UGC and AIGC, such as diverse real-world content and sophisticated generative enhancements, it is plausible that existing perceptual IQA models could be directly applicable to AI-UGC quality assessment. However, as illustrated in Fig. ~\ref{fig:First_Image}, due to the lack of dedicated datasets specifically targeting AI-enhanced user-generated content (AI-UGC), it remains unclear how well current models generalize to AI-UGC scenarios, and whether they can reliably assess the perceptual quality of such content. To address this gap, we introduce the \textbf{AU-IQA} dataset, as shown in Fig. ~\ref{fig_overall}, a dedicated benchmark specifically designed for perceptual quality assessment of AI-enhanced user-generated content. On this database, we can further explore the performance of existing IQA models on AI-UGC scenarios, identify limitations about quality assessment on AI-UGC, and  fosters the development of AI-UGC perceptual quality assessment methods.  The main contributions of our work include:

\begin{itemize}[leftmargin=*]
\item A clear definition and scope for AI-enhanced content have not yet to be established. This has greatly impacted the development of AI-enhanced models and user experience. In order to solve this problem, we propose the concept and background of \textbf{AI-UGC}. 
\item We present \textbf{AU-IQA} (\textbf{A}I models enhanced \textbf{U}ser-generated \textbf{I}mage \textbf{Q}uality \textbf{A}ssessment), an AI-UGC quality evaluation dataset with Mean Opinion Score (MOS), comprising 4.8k AI-UGC images enhanced by three types of AI techniques, including super-resolution, low-light enhancement and denoising. Then we test existing UGC and AIGC quality assessment models on our proposed dataset.
\item We benchmark a range of existing perceptual IQA models on AU-IQA, systematically analyzing their effectiveness.

\end{itemize}

\section{RELATED WORK}

\subsection{AI-based Enhancement of User-Generated Content}
Apart from direct image generation, recent years have witnessed rapid advancements in AI-based enhancement techniques for user-generated content (UGC) ~\cite{li2025imageenhancement}, including super-resolution ~\cite{lin2023diffbir, wu2024one, yang2023pasd, Yu_2024_CVPR}, denoising ~\cite{Chen_2023_CVPR} and low-light enhancement ~\cite{Jiang_2024_ECCV, Han_ECCV24_GLARE, jiang2025dark, FastLLVE, Light-VQA+}. Deep learning models have been widely adopted to improve the perceptual quality of UGC, with various techniques emerging over time. Early approaches were dominated by convolutional neural networks (CNNs), such as DnCNN ~\cite{zhang2017beyond} for image denoising and SRCNN ~\cite{dong2016image} for super-resolution, which demonstrated strong performance on traditional degradation types. Subsequently, generative adversarial networks (GANs) were introduced to further enhance perceptual quality, with notable methods like ESRGAN ~\cite{wang2018esrgan} significantly improving texture recovery and realism. More recently, transformer-based architectures, such as Restormer ~\cite{zamir2022restormer} and Uformer ~\cite{wang2022uformer}, have shown superior capability in handling complex distortions through long-range dependency modeling. Furthermore, diffusion models have started to gain attention for their ability to generate high-fidelity enhancements with controllable quality, as demonstrated by methods like IR-SDE ~\cite{luo2023refusion} for image restoration and Palette ~\cite{saharia2022palette} for image-to-image enhancement, marking a new frontier in AI-based UGC enhancement.

As AI-enhanced models become more widely used, it is important for quality evaluation methods to keep up. Developing reliable quality assessment dataset and accurate quality assessment models will help better measure the performance of these AI-enhanced techniques and further improve their results, leading to more realistic and visually appealing content.


\subsection{Image Quality Assessment}
Image Quality Assessment (IQA) refers to the task of evaluating the quality of images in a way that aligns with human visual perception. Traditional full-reference IQA metrics, such as PSNR and SSIM, assess image quality by comparing pixel differences between the reference and distorted images. While these metrics are widely used, they primarily focus on low-level pixel fidelity and often fail to capture higher-level perceptual features that align more closely with human subjective evaluations.

Recent developments in image quality assessment ~\cite{zhang2023perceptualqualityassessmentexploration} have seen the emergence of learning-based no-reference models like NIMA ~\cite{talebi2018nima}, DBCNN ~\cite{zhang2019dbcnn}, and HyperIQA ~\cite{su2020hyperiqa} for UGC, followed by models  which use visual-textual alignment to improve assessment accuracy such as CLIP-IQA ~\cite{wang2022exploring}, MUSIQ ~\cite{ke2021musiq} and LIQE ~\cite{zhang2023blind}. The latest, such as Q-Align ~\cite{wu2023qalign} and Q-Eval-Score ~\cite{zhang2025qeval} utilize large multimodal models (LMMs) ~\cite{surveyaqualitylmm} and proposes a discrete-level-based training strategy to better align quality prediction with human perception. As for datasets, UGC-related ones include TID2013 ~\cite{niko2013tid} and KADID-10K ~\cite{lin2019kadid}, while AIGC-focused datasets such as AGIQA-3k ~\cite{li2023agiqa}, Pick-A-Pic ~\cite{kirstain2023pickapicopendatasetuser} and HPS v2 ~\cite{wu2023human} are designed to assess AI-generated content. However, existing studies primarily focus on either UGC or purely AIGC scenarios, and there is still a lack of dedicated models and datasets targeting the perceptual quality of AI-generated user content (AI-UGC). To bridge this gap, we propose the dataset \textbf{AU-IQA} for AI-UGC quality assessment.

\begin{figure}[htbp]
\hspace{-0.1cm}
\centerline{\includegraphics[scale=0.40]{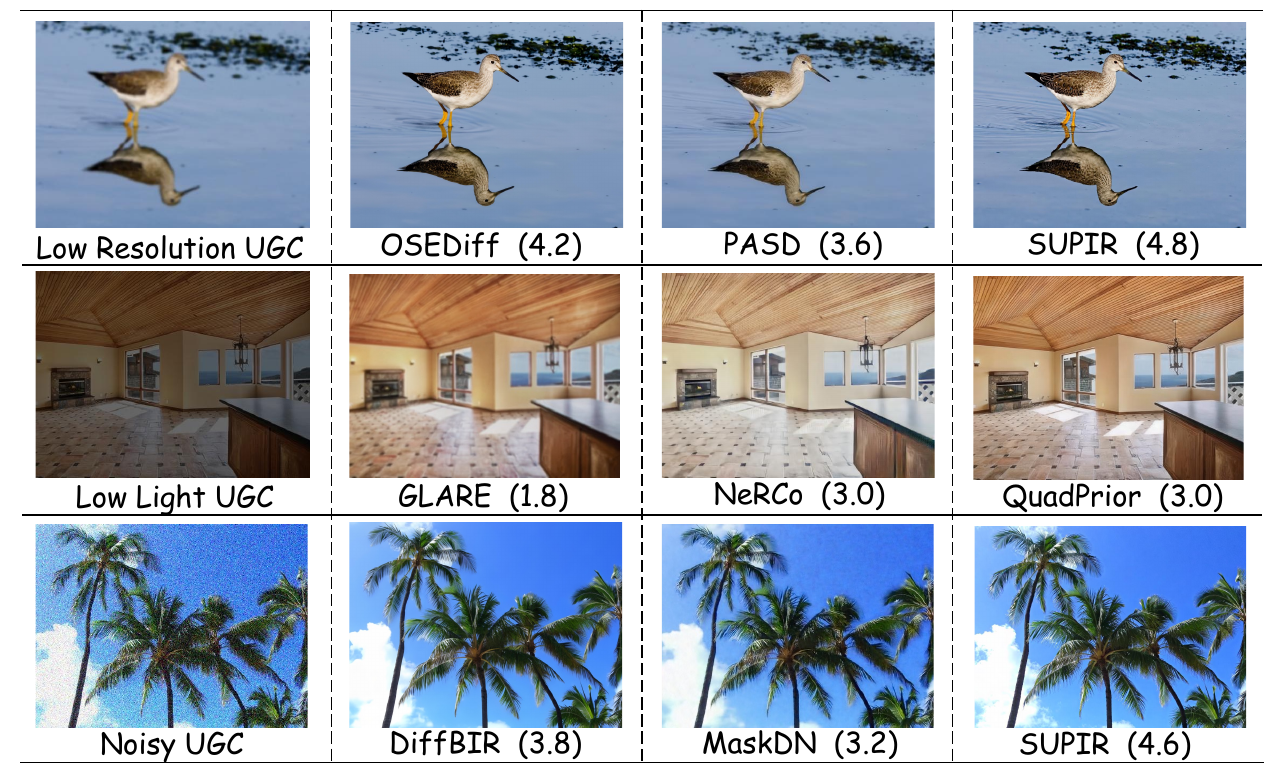}}
\caption{The first col of images are UGC with various quality issues, others are AI-UGC generated by different models along with their corresponding MOSs. }
\vspace{-0.6cm} %
\label{fig_annotation}
\end{figure}

\begin{figure*}[!t]
\captionsetup{aboveskip=5pt, belowskip=0pt} 
\hspace{2mm}
\centerline{\hspace*{-0.35cm}\includegraphics[scale=0.47]{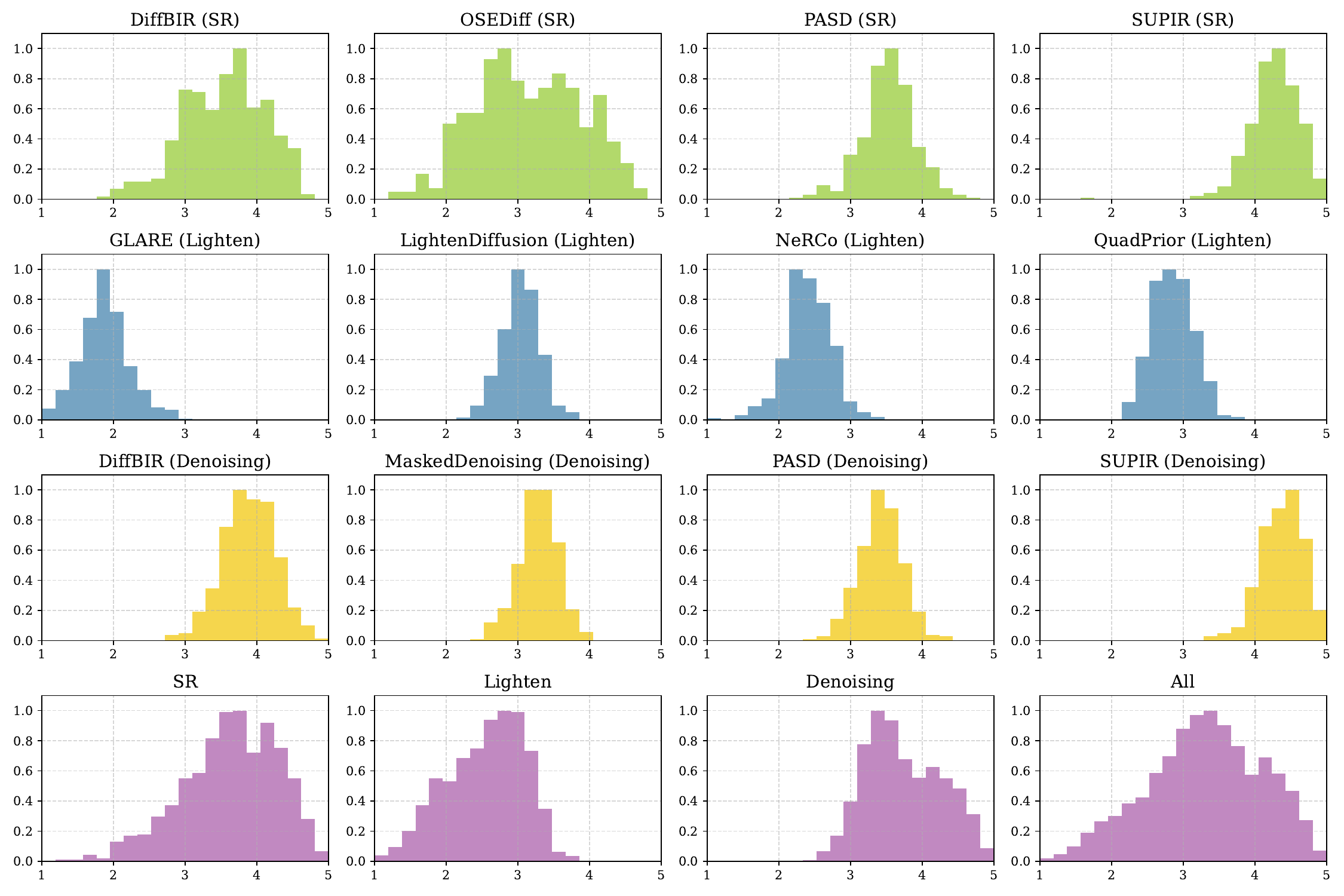}} 
\caption{Distribution of MOS Scores for Different Degradation Types and Enhancement Models.}
\vspace{-0.cm} %
\label{fig_distribution}
\end{figure*}

\section{Dataset Construction}

\subsection{AI-based Enhancement Models}
To build a comprehensive AI-UGC dataset, we first introduce three typical degradation types commonly encountered in real-world user-generated content: \textbf{Low Resolution (LR)}, \textbf{Low Light} and \textbf{Noise Corruption}. For each type, four representative enhancement models are selected. Specifically, for super-resolution, we adopt DiffBIR ~\cite{lin2023diffbir}, OSEDiff ~\cite{wu2024one}, PASD ~\cite{yang2023pasd}, and SUPIR ~\cite{Yu_2024_CVPR}. For low-light enhancement, we employ GLARE ~\cite{Han_ECCV24_GLARE}, LightenDiffusion ~\cite{Jiang_2024_ECCV}, NeRCo ~\cite{Yang_2023_ICCV}, and QuadPrior ~\cite{Wang_2024_CVPR}. For denoising, we apply DiffBIR, MaskedDenoising ~\cite{Chen_2023_CVPR}, PASD, and SUPIR. These models, spanning a diverse range of architectures, ensure the generality and diversity of the enhanced UGC content.

\subsection{Data Collection and Annotation}
Constructing a large-scale AI-UGC dataset poses significant challenges, as real-world UGC images often exhibit uncontrollable quality variations and lack standardized degradation types. To enable systematic evaluation and controlled analysis, we adopt a synthetic degradation strategy based on high-quality UGC images. Specifically, we select 400 high-quality samples from the KonIQ-10k dataset, based on their subjective quality scores. Three common degradation types—low resolution, low-light conditions, and noise—are then applied to these images through controlled manipulations, including downsampling, brightness reduction, and Gaussian noise addition. Each degraded image is subsequently restored using four different enhancement models corresponding to each degradation type, as introduced in Section 3.1, resulting in a total of $3 \times 4 \times 400 = 4,800$ enhanced images.

For annotation, we adopt the Mean Opinion Score (MOS) approach. Each image is independently rated by five trained annotators following a unified guideline. Annotators assign a quality score from 1 to 5, where higher scores indicate better perceptual quality. Evaluation focuses on four aspects—resolution, brightness, noise, and distortion—with resolution and distortion receiving greater attention. The annotation starts with an initial score of 5, deducting points according to the severity of observed degradations. An illustrative example of the collected data and corresponding annotations is shown in Fig. ~\ref{fig_annotation}. To ensure the reliability and consistency of the annotations, The annotations are required cross-checked for consistency. In cases where significant discrepancies are found, the respective annotators are requested to revisit and relabel the samples. After verifying annotation consistency, the final MOS is calculated as the average of all annotators’ scores.

\begin{table*}[!t]
\centering
\caption{Performance of quality assessment models comparison on the each type of AI-UGC. [\textbf{\textcolor{red}{red}}: best in each group, \textbf{\textcolor{red}{\uline{\textcolor{red}{red}}}}: best in all, PLCC: $\uparrow$, SRCC: $\uparrow$]}
\renewcommand{\arraystretch}{1.2}
\setlength{\tabcolsep}{0.6pt} 
\renewcommand{\arraystretch}{1.3} 
\scalebox{0.85}{
\begin{tabular}{l|cc|cc|cc|cc|cc|cc|cc|cc|cc|cc|cc|cc}
\toprule[0.5mm]
Type & 
\multicolumn{8}{c|}{\textbf{SR}} & 
\multicolumn{8}{c|}{\textbf{LL Enhancement}} & 
\multicolumn{8}{c}{\textbf{Denoising}} \\ 
\cline{2-25}
& 
\multicolumn{2}{c|}{DiffBIR} & 
\multicolumn{2}{c|}{OSEDiff} & 
\multicolumn{2}{c|}{PASD} & 
\multicolumn{2}{c|}{SUPIR} & 
\multicolumn{2}{c|}{GLARE} & 
\multicolumn{2}{c|}{LightDiff} & 
\multicolumn{2}{c|}{NeRCo} & 
\multicolumn{2}{c|}{QuadPrior} & 
\multicolumn{2}{c|}{DiffBIR} & 
\multicolumn{2}{c|}{MaskDN} & 
\multicolumn{2}{c|}{PASD} & 
\multicolumn{2}{c}{SUPIR} \\ 
\cline{1-25}
Corr & PLCC & SRCC & PLCC & SRCC & PLCC & SRCC & PLCC & SRCC & PLCC & SRCC & PLCC & SRCC & PLCC & SRCC
& PLCC & SRCC & PLCC & SRCC & PLCC & SRCC & PLCC & SRCC & PLCC & SRCC\\ 
\hline
\multicolumn{25}{c}{\textbf{UGC Image Quality Assessment}} \\ 
\hline
\raggedright TOPIQ & 0.164 & 0.156 & 0.418 & 0.384 & 0.457 & 0.412 & 0.246 & 0.126 & 0.648 & 0.611 & 0.403 & 0.379 & 0.387 & 0.377 & 0.359 & 0.341 & 0.173 & 0.148 & 0.328 & 0.328 & 0.204 & 0.220 & 0.256 & 0.194 \\ 
\raggedright LIQE & 0.260 & 0.271 & 0.576 & 0.567 & 0.395 & 0.454 & 0.217 & 0.185 & 0.553 & 0.467 & 0.408 & 0.387 & 0.474 & 0.482 & \textbf{\textcolor{red}{\uline{\textcolor{red}{0.398}}}} & \textbf{\textcolor{red}{\uline{\textcolor{red}{0.397}}}} & 0.143 & 0.127 & 0.275 & 0.269 & 0.059 & 0.253 & 0.260 & \textbf{\textcolor{red}{\uline{\textcolor{red}{0.278}}}} \\ 
\raggedright ARNIQA & -0.074 & -0.092 & 0.086 & 0.081 & 0.103 & 0.102 & 0.199 & 0.179 & 0.587 & 0.573 & 0.255 & 0.243 & 0.290 & 0.290 & 0.274 & 0.300 & -0.108 & -0.086 & 0.142 & 0.110 & 0.155 & 0.164 & 0.086 & 0.106 \\ 
\raggedright HyperIQA & 0.382 & 0.353 & 0.619 & 0.626 & 0.472 & 0.455 & \textbf{\textcolor{red}{\uline{\textcolor{red}{0.347}}}} & \textbf{\textcolor{red}{\uline{\textcolor{red}{0.211}}}} & 0.234 & 0.217 & 0.390 & 0.363 & 0.287 & 0.316 & 0.316 & 0.315 & 0.228 & 0.207 & 0.333 & 0.308 & 0.248 & 0.285 & 0.218 & 0.169 \\ 
\raggedright InternVL2 & 0.288 & 0.289 & 0.394 & 0.363 & 0.367 & 0.382 & 0.230 & 0.158 & 0.641 & 0.650 & 0.349 & 0.330 & 0.188 & 0.192 & 0.245 & 0.250 & 0.113 & 0.106 & 0.156 & 0.158 & 0.198 & 0.234 & 0.155 & 0.109 \\ 
\raggedright Qwen2-VL & 0.292 & 0.275 & 0.508 & 0.428 & 0.255 & 0.234 & 0.122 & 0.100 & 0.605 & 0.580 & 0.278 & 0.267 & 0.316 & 0.303 & 0.174 & 0.183 & 0.129 & 0.149 & 0.232 & 0.260 & 0.100 & 0.161 & 0.068 & 0.115 \\ 
\raggedright LLaVA-v1.6 &0.264  & 0.284 & 0.549 & 0.545 & 0.380 & 0.355 & 0.145 & 0.100 & 0.614 & 0.608 & 0.418 & 0.409 & 0.478 & 0.446 & 0.271 & 0.261 & 0.047 & 0.067 & 0.254 & 0.275 & 0.136 & 0.166 & 0.214 & 0.238 \\ 
\raggedright Q-Align & 0.433 & 0.429 & 0.702 & 0.696 & \textbf{\textcolor{red}{\uline{\textcolor{red}{0.479}}}} & \textbf{\textcolor{red}{\uline{\textcolor{red}{0.475}}}} & 0.345 & 0.190 & \textbf{\textcolor{red}{\uline{\textcolor{red}{0.731}}}} & \textbf{\textcolor{red}{\uline{\textcolor{red}{0.719}}}} & 0.474 & 0.440 & \textbf{\textcolor{red}{\uline{\textcolor{red}{0.551}}}} & \textbf{\textcolor{red}{\uline{\textcolor{red}{0.492}}}} & 0.372 & 0.394 & \textbf{\textcolor{red}{\uline{\textcolor{red}{0.277}}}} & 0.249 & \textbf{\textcolor{red}{\uline{\textcolor{red}{0.346}}}} & \textbf{\textcolor{red}{\uline{\textcolor{red}{0.340}}}} & \textbf{\textcolor{red}{\uline{\textcolor{red}{0.378}}}} & \textbf{\textcolor{red}{\uline{\textcolor{red}{0.368}}}} & \textbf{\textcolor{red}{\uline{\textcolor{red}{0.299}}}} & 0.273 \\
\raggedright Qwen2-VL * &0.342  & 0.357 & 0.521 & 0.467 & 0.289 & 0.167 & 0.254 & 0.163 & 0.587 & 0.492 & 0.368 & 0.356 & 0.401 & 0.364 & 0.222 & 0.236 & 0.258 & 0.227 & 0.282 & 0.263 & 0.100 & 0.107 & 0.235 & 0.209 \\  
\raggedright DEQA &\textbf{\textcolor{red}{\uline{\textcolor{red}{0.444}}}}  & \textbf{\textcolor{red}{\uline{\textcolor{red}{0.433}}}} & \textbf{\textcolor{red}{\uline{\textcolor{red}{0.703}}}} & \textbf{\textcolor{red}{\uline{\textcolor{red}{0.733}}}} & 0.467 & 0.462 & 0.312 & 0.136 & 0.726 & 0.707 & \textbf{\textcolor{red}{\uline{\textcolor{red}{0.498}}}} & \textbf{\textcolor{red}{\uline{\textcolor{red}{0.483}}}} & 0.489 & 0.456 & 0.335 & 0.356 & 0.262 & \textbf{\textcolor{red}{\uline{\textcolor{red}{0.250}}}} & 0.332 & 0.310 & 0.268 & 0.297 & 0.245 & 0.178 \\  

\hline
\multicolumn{25}{c}{\textbf{AIGC Image Quality Assessment}} \\ 
\hline
\raggedright MA-AGIQA & 0.084 & 0.090 & 0.321 & 0.311 & 0.008 & 0.005 & -0.021 & 0.009 & 0.398 & 0.375 & 0.121 & 0.100 & 0.070 & 0.035 & -0.006 & -0.008 & -0.075 & -0.075 & 0.069 & 0.071 & 0.025 & 0.045 & 0.020 & 0.030 \\ 
\raggedright InternVL2 & \textbf{\textcolor{red}{0.333}} & \textbf{\textcolor{red}{0.317}} & 0.281 & 0.251 & \textbf{\textcolor{red}{0.284}} & \textbf{\textcolor{red}{0.264}} & 0.213 & 0.147 & 0.338 & 0.330 & \textbf{\textcolor{red}{0.140}} & 0.122 & 0.273 & 0.306 & 0.074 & 0.042 & 0.058 & 0.069 & 0.060 & 0.049 & 0.137 & 0.149 & \textbf{\textcolor{red}{0.213}} & \textbf{\textcolor{red}{0.174}} \\ 
\raggedright Qwen2-VL & 0.261 & 0.238 & 0.348 & 0.314 & 0.215 & 0.184 & \textbf{\textcolor{red}{0.244}} & \textbf{\textcolor{red}{0.206}} & \textbf{\textcolor{red}{0.400}} & \textbf{\textcolor{red}{0.423}} & 0.135 & 0.148 & 0.300 & \textbf{\textcolor{red}{0.332}} & 0.050 & 0.049 & \textbf{\textcolor{red}{0.100}} & \textbf{\textcolor{red}{0.090}} & 0.074 & 0.087 & 0.105 & 0.114 & 0.154 & 0.122 \\ 
\raggedright LLaVA-v1.6 & 0.272 & 0.256 & 0.362 & \textbf{\textcolor{red}{0.337}} & 0.225 & 0.202 & 0.140 & 0.133 & 0.274 & 0.271 & 0.138 & 0.109 & \textbf{\textcolor{red}{0.335}} & 0.319 & \textbf{\textcolor{red}{0.083}} & \textbf{\textcolor{red}{0.101}} & 0.033 & 0.014 & 0.076 & 0.049 & \textbf{\textcolor{red}{0.176}} & \textbf{\textcolor{red}{0.179}} & 0.180 & 0.139 \\ 
\raggedright Q-Align &0.229  & 0.248 & \textbf{\textcolor{red}{0.363}} & 0.335 & 0.178 & 0.189 & 0.058 & 0.073 & 0.213 & 0.198 & 0.087 & 0.074 & 0.284 & 0.281 & 0.051 & 0.042 & 0.040 & 0.017 & \textbf{\textcolor{red}{0.138}} & \textbf{\textcolor{red}{0.143}} & 0.118 & 0.141 & 0.109 & 0.103 \\ 
\raggedright Qwen2-VL * &0.225  & 0.234 & 0.352 & 0.252 & 0.157 & 0.121 & 0.100 & 0.059 & 0.420 & 0.397 & 0.128 & \textbf{\textcolor{red}{0.154}} & 0.329 & 0.312 & 0.079 & 0.081 & 0.039 & 0.027 & 0.048 & 0.057 & 0.018 & 0.036 & 0.132 & 0.107 \\

\raggedright Q-Eval-Score &0.283  & 0.297 & 0.161 & 0.134 & 0.178 & 0.205 & 0.192 & 0.217 & 0.224 & 0.208 & 0.191 & 0.173 & 0.257 & 0.250 & 0.181 & 0.181 & 0.204 & 0.231 & 0.162 & 0.195 & 0.235 & 0.258 & 0.213 & 0.216 \\ 
\hline
\multicolumn{25}{c}{\textbf{UGC and AIGC Image Quality Assessment}} \\ 
\hline
\raggedright InternVL2 & 0.324 & 0.334 & 0.433 & 0.383 & 0.357 & 0.358 & \textbf{\textcolor{red}{0.253}} & \textbf{\textcolor{red}{0.178}} & 0.619 & 0.580 & 0.333 & 0.317 & 0.190 & 0.213 & 0.190 & 0.193 & 0.127 & 0.127 & 0.233 & 0.227 & 0.092 & 0.135 & 0.243 & 0.219 \\ 
\raggedright Qwen2-VL & 0.254 & 0.238 & 0.461 & 0.379 & 0.200 & 0.189 & 0.109 & 0.098 & 0.545 & 0.509 & 0.179 & 0.155 & 0.225 & 0.189 & 0.157 & 0.168 & 0.118 & 0.105 & 0.188 & 0.162 & 0.030 & 0.086 & 0.079 & 0.128 \\ 
\raggedright LLaVA-v1.6 & 0.260 & 0.268 & 0.453 & 0.422 & \textbf{\textcolor{red}{0.383}} & \textbf{\textcolor{red}{0.363}} & 0.181 & 0.127 & 0.660 & 0.657 & \textbf{\textcolor{red}{0.350}} & \textbf{\textcolor{red}{0.322}} & 0.260 & 0.302 &\textbf{\textcolor{red}{0.280}} &0.247 &0.081 &0.095 &0.213 &0.242 &\textbf{\textcolor{red}{0.145}} &\textbf{\textcolor{red}{0.177}} &0.220 &0.220 \\
\raggedright Q-Align &0.242  & 0.272 & 0.491 & \textbf{\textcolor{red}{0.476}} & 0.326 & 0.277 & 0.228 & 0.073 & \textbf{\textcolor{red}{0.661}} & \textbf{\textcolor{red}{0.664}} & 0.301 & 0.291 & \textbf{\textcolor{red}{0.450}} & 0.383 & 0.266 & \textbf{\textcolor{red}{0.287}} & 0.035 & 0.013 & \textbf{\textcolor{red}{0.306}} & \textbf{\textcolor{red}{0.287}} & 0.103 & 0.060 & 0.125 & 0.079 \\ 
\raggedright Qwen2-VL * & \textbf{\textcolor{red}{0.372}}  & \textbf{\textcolor{red}{0.362}} & \textbf{\textcolor{red}{0.529}} & 0.472 & 0.243 & 0.190 & 0.247 & 0.150 & 0.570 & 0.543 & 0.307 & 0.308 & 0.434 & \textbf{\textcolor{red}{0.402}} & 0.229 & 0.236 & \textbf{\textcolor{red}{0.222}} & \textbf{\textcolor{red}{0.189}} & 0.232 & 0.223 & 0.028 & 0.063 & \textbf{\textcolor{red}{0.252}} & \textbf{\textcolor{red}{0.225}} \\ 
\bottomrule[0.5mm] 
\end{tabular}
}
\label{tab:quality_assessment_model}
\end{table*}

\subsection{Statistical Analysis}

The histograms of the MOS for each AI-based enhancement model and type are shown in Fig.~\ref{fig_distribution}. For all three types, the MOS distributions of enhancement models display distinct patterns. \textbf{For super resolution (SR)}, DiffBIR and OSEDiff show relatively uniform MOS distributions, while PASD clusters around 3.5 and SUPIR is concentrated above 4, indicating superior performance. \textbf{For low light enhancement (LLE)}, all four models yield lower MOS scores, with distributions resembling a normal curve; GLARE centers between 1.5 and 2.5, LightenDiffusion peaks around 3, NeRCo between 2 and 3, and QuadPrior between 2.5 and 3.5. \textbf{For denoising}, the MOS distributions are also approximately normal but shift higher, with most scores between 3.5 and 4.5.

The overall MOS distributions across the three AI enhancement types, super resolution, low light enhancement, and denoising highlight the diversity of our AI-UGC dataset. As shown in the histograms, each AI enhancement type exhibits distinct characteristics, reflecting the challenges faced by IQA models in evaluating AI-enhanced content. The MOS distributions for types including super resolution, low light enhancement, and denoising highlight the diversity of AI-enhanced images. For super resolution, the MOS scores range from 2 to 4.5, reflecting varying levels of super resolution enhancement. Scores for low light enhancement are mostly between 1.5 and 3.5, reflecting the difficulties involved in enhancing images captured under low-light conditions. Denoised images achieve MOS scores concentrated between 3.5 and 4.5, suggesting more stable and reliable denoising performance. The diversity of these distributions reflects the realistic nature of AI-enhanced content and ensures the dataset's suitability for evaluating the effectiveness of IQA models across different degradation conditions.

\begin{table}[!tbp]
\centering
\caption{Performance of quality assessment models comparison on AI-UGC. The random refers to randomly selecting 1600 images from the dataset. The all refers to the whole dataset. [\textbf{\textcolor{red}{red}}: best in each group, \textbf{\textcolor{red}{\uline{\textcolor{red}{red}}}}: best in all].}
\renewcommand{\arraystretch}{1.2}
\setlength{\tabcolsep}{0.8pt} 
\renewcommand{\arraystretch}{1.3} 
\resizebox{\columnwidth}{!}{ 
\begin{tabular}{l|cc|cc|cc|cc|cc}
\toprule[0.5mm] 
\multicolumn{1}{l|}{Type} &
\multicolumn{2}{c|}{\makecell{\textbf{SR}}} & 
\multicolumn{2}{c|}{\makecell{\textbf{LLE}}} & 
\multicolumn{2}{c|}{\makecell{\textbf{Denoising}}} & 
\multicolumn{2}{c|}{\makecell{\textbf{Random}}} & 
\multicolumn{2}{c}{\makecell{\textbf{All}}} \\ 
\cline{1-11}
\multicolumn{1}{l|}{Corr} & PLCC & SRCC & PLCC & SRCC & PLCC & SRCC & PLCC & SRCC & PLCC & SRCC\\ 
\hline
\multicolumn{11}{c}{\textbf{UGC Image Quality Assessment}} \\ 
\hline
\raggedright TOPIQ & 0.541 & 0.501 & 0.770 & 0.694 & 0.120 & 0.122  & 0.725 & 0.668 & 0.721 & 0.661\\ 
\raggedright LIQE & 0.564 & 0.505 & 0.789 & 0.742 & 0.051 & -0.004  & 0.648 & 0.548 & 0.650 & 0.549\\ 
\raggedright ARNIQA & 0.219 & 0.190 & 0.762 & 0.714 & -0.260 & -0.284 & 0.429 & 0.281 & 0.416 & 0.269\\ 
\raggedright hyperIQA & 0.410 & 0.439 & 0.648 & 0.542 & \textbf{\textcolor{red}{\uline{\textcolor{red}{0.684}}}} & \textbf{\textcolor{red}{\uline{\textcolor{red}{0.697}}}}  & 0.720 & 0.694 & 0.723 & 0.695\\ 
\raggedright InternVL2 & 0.342 & 0.350 & 0.660 & 0.479 & 0.574 & 0.610  & 0.670 & 0.603 & 0.671 & 0.607\\ 
\raggedright Qwen2-VL & 0.172 & 0.129 & 0.671 & 0.658 & 0.183 & 0.273  & 0.599 & 0.539 & 0.597 & 0.536\\ 
\raggedright LLaVA-v1.6 & 0.405 & 0.374 & 0.763 & 0.739 & 0.458 & 0.460  & 0.723 & 0.664 & 0.721 & 0.659\\ 
\raggedright Q-Align & 0.551 & 0.480 & 0.815 & 0.751 & 0.526 & 0.522  & \textbf{\textcolor{red}{\uline{\textcolor{red}{0.795}}}}& \textbf{\textcolor{red}{\uline{\textcolor{red}{0.777}}}}& \textbf{\textcolor{red}{\uline{\textcolor{red}{0.796}}}} & \textbf{\textcolor{red}{\uline{\textcolor{red}{0.771}}}}\\ 

\raggedright Qwen2-VL * & 0.417 & 0.366 & 0.771 & 0.701 & 0.556 & 0.606  & 0.732 & 0.687 & 0.720 & 0.673\\
\raggedright DEQA & \textbf{\textcolor{red}{\uline{\textcolor{red}{0.587}}}} & \textbf{\textcolor{red}{\uline{\textcolor{red}{0.523}}}} & \textbf{\textcolor{red}{\uline{\textcolor{red}{0.832}}}} & \textbf{\textcolor{red}{\uline{\textcolor{red}{0.791}}}} & 0.383 & 0.385  & 0.767 & 0.746 & 0.766 & 0.749\\ 
\hline
\multicolumn{11}{c}{\textbf{AIGC Image Quality Assessment}} \\ 
\hline
\raggedright MA-AGIQA & 0.159 & 0.148 & 0.473 & 0.418 & 0.114 & 0.120   & 0.429 & 0.281 & 0.374 & 0.321\\ 
\raggedright InternVL2 & 0.243 & 0.266 & 0.395 & 0.353 & \textbf{\textcolor{red}{0.591} }& \textbf{\textcolor{red}{0.598}}   & \textbf{\textcolor{red}{0.602}} &\textbf{\textcolor{red}{0.596}} & \textbf{\textcolor{red}{0.606}} & \textbf{\textcolor{red}{0.596}}\\ 
\raggedright Qwen2-VL & 0.244 & 0.206 & \textbf{\textcolor{red}{0.562}} & \textbf{\textcolor{red}{0.528}} & 0.304 & 0.309   &0.570 & 0.510 & 0.557 & 0.500\\ 
\raggedright LLaVA-v1.6 & \textbf{\textcolor{red}{0.314}} & \textbf{\textcolor{red}{0.299}} & 0.459 & 0.447 & 0.400 & 0.417 & 0.583 &0.576 & 0.566 & 0.555\\ 
\raggedright Q-Align & 0.143 & 0.145 & 0.388 & 0.367 & 0.323 & 0.353  & 0.530 & 0.533 & 0.516 & 0.519\\
\raggedright Qwen2-VL * & 0.288 & 0.303 & 0.650 & 0.590 & 0.570 & 0.604  & 0.673 & 0.634 & 0.665 & 0.622\\

\raggedright Q-Eval-Score & 0.239 & 0.234 & 0.250 & 0.246 & 0.230 & 0.240  & 0.243 & 0.243 & 0.258 & 0.256\\
\hline
\multicolumn{11}{c}{\textbf{UGC and AIGC Image Quality Assessment}} \\ 
\hline
\raggedright InternVL2 & 0.305 & 0.298 & 0.652 & 0.398 & \textbf{\textcolor{red}{0.635}} & \textbf{\textcolor{red}{0.660}}   &  0.689 &0.617 & 0.683 & 0.616\\ 
\raggedright Qwen2-VL  & 0.090 & 0.122 & 0.664 & 0.530 & 0.231 & 0.330   & 0.589 &0.532 & 0.568 & 0.518\\ 
\raggedright LLaVA-v1.6 & 0.316 & 0.304 & 0.658 & 0.482 & 0.479 & 0.490   & 0.624 &0.504 & 0.624 & 0.518\\ 
\raggedright Q-Align & 0.298 & 0.269 & \textbf{\textcolor{red}{0.742}} & \textbf{\textcolor{red}{0.619}} & 0.329 & 0.402 & 0.674 & 0.635 & 0.672 & 0.632\\ 
\raggedright Qwen2-VL  & \textbf{\textcolor{red}{0.367}} & \textbf{\textcolor{red}{0.341}} & 0.736 & 0.665 & 0.598 & 0.656  &\textbf{\textcolor{red}{0.733}} & \textbf{\textcolor{red}{0.685}} & \textbf{\textcolor{red}{0.722}} & \textbf{\textcolor{red}{0.674}}\\
\bottomrule[0.5mm] 
\end{tabular}
}
\label{tab:quality_assessment_type}
\end{table}

\section{Experiments} 

\subsection{Benchmark Models}
AI-UGC exhibit characteristics of both UGC and AIGC. To effectively evaluate the capability of current models in assessing such hybrid content and lay the foundation for future research on AI-UGC quality assessment, we select representative models from three categories for benchmarking: UGC IQA models, AIGC IQA models, and large multimodal models (LMMs) that have demonstrated strong performance in image-text understanding tasks. In our experiments, we finetune LMMs on KADID-10K ~\cite{lin2019kadid} to serve as UGC quality assessment models, and on AGIQA-3K ~\cite{li2023agiqa} to serve as AIGC quality assessment models. These models serve as a comprehensive baseline for analyzing the strengths and limitations of existing approaches in the context of AI-UGC image quality assessment. The selected models can be categorized into three groups:

\begin{itemize}[leftmargin=*]
\item Traditional image quality assessment models: This group includes TOPIQ ~\cite{chen2024topiq}, LIQE ~ \cite{zhang2023blind}, ARNIQA ~ \cite{agnolucci2024arniqa}, and hyperIQA ~ \cite{Su_2020_CVPR}. Thess four models are UGC quality assessment models. For the AIGC quality assessment model, since most existing models use the SBS evaluation format, we only selected one model MA-AGIQA\cite{wang2024large}.
\item Large Multimodal Models: This group includes three outstanding LMMs: InternVL2-8B ~ \cite{chen2024internvl}, Qwen2-VL-7B-Instruct ~ \cite{qwen} and LLaVA-v1.6-mistral-7B ~ \cite{liu2024llavanext}.  Considering the strong task adaptation capabilities of LMMs through fine-tuning, we fine-tune these models to directly output quality scores using the UGC image quality assessment dataset, the AIGC image quality assessment dataset, or a combination of both.
\item LMM-based models for quality assessment: This group includes models Q-Align ~\cite{wu2023qalign}, Qwen2-VL-7B-Instruct* ~ \cite{qwen} (using the fine-tuning strategy of Q-Align), Q-Eval-Score ~ \cite{zhang2025qeval}, and DeQA ~ \cite{you2025deqa_score}, which adapt large multimodal models for the task of image quality assessment. Unlike standard LMMs that directly output scalar quality scores, some of these models are trained to predict discrete quality ratings, which are then mapped to numerical values. This approach enables more stable and interpretable assessments. 

\end{itemize}

\subsection{Performance Discussion}
The performance on our dataset is presented in Table ~\ref{tab:quality_assessment_model} and Table ~\ref{tab:quality_assessment_type},  with the evaluation metrics including Pearson Linear Correlation Coefficient (PLCC) and Spearman Rank Correlation Coefficient (SRCC), which are standard metrics widely used for assessing image quality. From these results,  which several key insights can be drawn:
\begin{itemize}[leftmargin=*]
    \item \textbf{Performance Comparison in UGC, AIGC, and Hybrid Models:} The UGC quality assessment model performs the best overall. Models capable of assessing both UGC and AIGC images rank second, while models designed specifically for AIGC quality evaluation perform the worst. This suggests that the features of AI-UGC images are more similar to those of traditional UGC images than to AIGC images.
    \item \textbf{Variation Across Different AI-UGC Types:} The performance of the same quality assessment model varies depending on the type of AI-UGC. For example,  HyperIQA achieves the best performance on super-resolution enhanced images but performs less well on low-light enhanced images. Overall, the quality assessment of low-light enhanced images appears to be more consistent across two other enhancement types. This indicates that existing quality assessment models exhibit varying effectiveness when applied to different types of AI-UGC, highlighting the need for more specialized approaches tailored to specific content.
    \item \textbf{Performance on Larger Datasets:} Models perform significantly better when evaluated on a larger-scale dataset, such as the entire AU-IQA dataset. This suggests that the performance of existing quality assessment models on AI-UGC is unstable, and the observed improvements on larger-scale test sets may result from the increased data volume making it easier for models to fit underlying linear relationships.

\end{itemize}
In summary, the AI-UGC quality assessment task is more aligned with traditional UGC quality evaluation, possibly because AI-UGC appears visually closer to UGC in terms of perceptual characteristics. However, due to the varying performance of the same model on different types of AI-UGC, while different types of AI-UGC may appear similar to the human eye, they exhibit significant feature differences when assessed by existing quality models. In addition, the performance of models on small-scale datasets is often unstable. Therefore, future research in AI-UGC quality assessment should take these differences into account and consider the unique quality characteristics of each type of AI-enhanced content.

\section{Conclusion} 
The application of AI-UGC is rapidly expanding, yet there is currently no dedicated quality assessment model for these. This gap hampers both user experience and the advancement of AI-UGC technologies. Given that AI-UGC exhibits characteristics of both UGC and AIGC, this paper explores the use of existing quality assessment models and large multimodal models for AI-UGC quality evaluation. We introduce AU-IQA, a benchmark dataset which includes 4.8K AI-enhanced images across three types of AI enhancement for AI-UGC quality assessment, and present a comprehensive evaluation of the performance of current quality assessment models, including traditional models and large multimodal models. The results indicate that UGC-focused models perform the best, achieving satisfactory results on a large-scale test dataset. This suggests that the quality assessment of AI-UGC is more similar to that of UGC. Meanwhile, the result of noticeable performance differences across different test points highlights the limitations and instability of existing models in AI-UGC quality assessment and aim to provide insights for future research in this area.
\begin{acks}
The work was supported in part by the National Natural Science Foundation of China under Grant 62301310, and in part by Sichuan Science and Technology Program under Grant 2024NSFSC1426.
\end{acks}

\bibliographystyle{ACM-Reference-Format}
\bibliography{sample-sigconf}


\end{document}